\documentclass[10pt]{article}

\usepackage{microtype}
\usepackage{booktabs}
\usepackage{caption}
\usepackage{url}
\usepackage{amsmath}
\usepackage{amssymb}
\usepackage{amsthm}

  \usepackage{geometry}
  \usepackage[hidelinks]{hyperref}
  \newenvironment{acknowledgements}{\section*{Acknowledgements}}{}

\newcommand{\cv}[1]{}
\newcommand{\av}[1]{#1}

\usepackage{natbib}
\usepackage{graphicx}
\usepackage{xcolor}
\usepackage{multirow}

\title{Algorithm Selection with Zero Domain Knowledge\\ via Text Embeddings\thanks{To appear in the Proceedings of AutoML 2026.}}

  \author{Stefan Szeider\\[4pt]
    \small Algorithms and Complexity Group\\[-3pt]
    \small TU Wien, Vienna, Austria\\[-3pt]
    \small \href{https://ac.tuwien.ac.at/szeider}{https://ac.tuwien.ac.at/szeider}
  }
  \date{}

\hypersetup{%
  pdftitle={Algorithm Selection with Zero Domain Knowledge via Text Embeddings},
  pdfauthor={Stefan Szeider},
  pdfsubject={Algorithm Selection, Text Embeddings, AutoML},
  pdfkeywords={algorithm selection, text embeddings, zero domain knowledge, ASlib, ZeroFolio}
}

\begin{document}

\maketitle
\av{\thispagestyle{empty}}

\begin{abstract} %
We propose ZeroFolio, a feature-free approach to algorithm selection that uses
pretrained text embeddings instead of hand-crafted instance features.
It reads the raw instance file as plain text, embeds it with a
pretrained embedding model, and selects an algorithm via weighted
$k$-nearest neighbors. Our approach is based on the observation that
pretrained embeddings can distinguish problem instances without any
domain knowledge or task-specific training. ZeroFolio applies to any problem
domain with text-based instance formats. We evaluate our
approach on 11 ASlib scenarios spanning 7 domains (SAT,
MaxSAT, QBF, ASP, CSP, MIP, and graph problems). ZeroFolio outperforms
a random forest trained on hand-crafted features in 9 of 11 scenarios,
often substantially, and in 8 of them with every serialization seed. It wins 8
of 11 scenarios against a per-scenario-tuned random forest. On the three
scenarios with published AutoFolio results from the 2015 ICON
Challenge, ZeroFolio comes within a small margin of AutoFolio without
any per-scenario tuning. Our ablation study on SAT12-ALL shows that
inverse-distance weighting and line shuffling improve performance. We further analyze the sensitivity of
our approach to the serialization seed. On the SAT12-ALL scenario, where
the random forest is stronger, both methods can be combined via soft
voting to achieve further improvements.
\end{abstract}

\section{Introduction} %
\label{sec:intro}

Algorithm selection is the fundamental problem of choosing a solver
from a portfolio of solvers for a given problem instance.
Fifty years ago, \citet{Rice1976} formalized this problem as a mapping from instance
features to algorithms, a formalization that has guided the
research since. Rice already saw the features as the hard part: ``The determination of the best
(or even good) features is one of the most important, yet nebulous, aspects of the
algorithm selection problem'' \citep[Sec.~2.2]{Rice1976}.

Current algorithm selection systems utilize hand-crafted features and
train machine learning models to predict solver performance.
SATzilla~\citep{Xu2008} pioneered feature-based algorithm selection for
propositional satisfiability (SAT) solvers. This approach was later
automated by AutoFolio~\citep{Lindauer2015}. The ASlib
benchmark~\citep{Bischl2016} provides a standardized evaluation framework
with precomputed features on various scenarios over several problem domains.

Standard algorithm selection systems hinge on domain-specific feature
engineering, since each domain requires expert knowledge to design
adequate features. Computing the features can itself be costly. For
instance, SATzilla's probing features run SAT solvers for a limited time.
\citet{Shavit2024} recently updated the SATzilla feature extractor and
report that the original tool failed to extract features from over a
fifth of modern SAT competition instances. Features designed for one
domain (e.g., clause-variable ratios for SAT) have to be re-implemented for
other domains (e.g., constraint satisfaction or answer set programming).
A practitioner faces a trade-off between investing in domain-specific
feature engineering and accepting a weaker algorithm selector.

We propose a new approach that does not require any feature
engineering. The basic idea is to read the raw instance file as plain
text, embed it with a pretrained off-the-shelf model,
and subsequently select an algorithm via weighted $k$-nearest
neighbors. Our results show that such embeddings produce
representations that are able to distinguish instances for
effective algorithm selection. This works without any task-specific
training or feature engineering.

It is not obvious that such a fixed pipeline generalizes across a
diverse set of formats, including
DIMACS CNF, weighted CNF, QDIMACS, MiniZinc models, MPS files, and
graph adjacency lists, without any domain-specific customization. However,
our experiments show that it does.

Our proposed approach uses line shuffling before the input text is
truncated, and we use weighted
$k$-nearest neighbors to select the algorithm. We evaluate ZeroFolio on
11 scenarios across 7 domains. Our approach outperforms a random forest
on hand-crafted features in 9 of the 11 scenarios (in 8 of them with every
serialization seed) and still wins in 8 of the 11 when the
random forest is tuned per scenario. Without any per-scenario tuning,
ZeroFolio comes within 6--9\% of AutoFolio on the three scenarios with
published results. Our ablation study on SAT12-ALL shows
that inverse-distance weighting and line shuffling improve performance.

\begin{figure}[t]
\centering
\includegraphics[width=\linewidth]{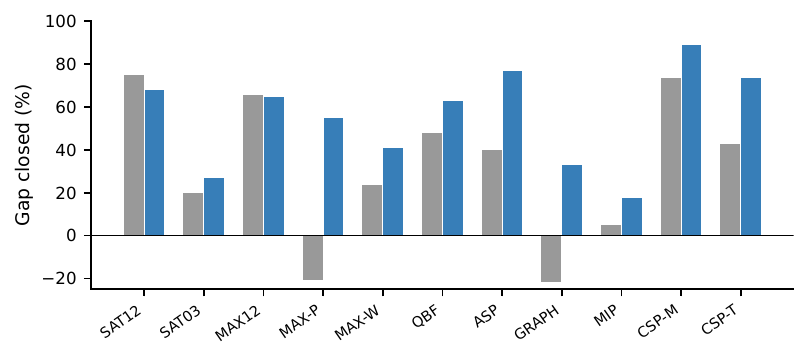}
\caption{Share of the gap between the single best solver (SBS) and the virtual best solver (VBS, best algorithm per instance) that a selector with PAR10 (penalized average runtime) $X$ closes: $100(\mathrm{SBS}-X)/(\mathrm{SBS}-\mathrm{VBS})$.
  \textcolor[HTML]{999999}{\rule{0.8em}{0.8em}}~random forest (RF) on hand-crafted features;\quad
  \textcolor[HTML]{377eb8}{\rule{0.8em}{0.8em}}~ZeroFolio (mean over
  seeds). RF falls below SBS on two scenarios.}
\label{fig:gap}
\end{figure}

Figure~\ref{fig:gap} gives a preview of the main result. On 9 of the
11 scenarios, ZeroFolio closes a larger fraction of the SBS--VBS gap
than the RF baseline, taking the mean over seeds.

\section{Related Work} %
\label{sec:related}

\paragraph{Classical algorithm selection.}
\citet{Rice1976} formalized algorithm selection as a mapping from
features to algorithms.
Many algorithm selection systems have been proposed~\citep{Kotthoff2016}.
SATzilla~\citep{Xu2008} instantiated this framework for SAT by
extracting structural features like clause-variable ratios, variable
interaction graphs, and probing statistics~\citep{Nudelman2004} and
training regression models to predict solver runtimes.
AutoFolio~\citep{Lindauer2015} further automated the configuration of the
selection pipeline.
The ASlib benchmark library~\citep{Bischl2016} provides the basis for a
standardized evaluation of algorithm selection systems across several domains.
The mentioned algorithm selection systems rely on hand-crafted,
domain-specific instance features. ASlib provides such features for all
of its scenarios.

\paragraph{Feature-free algorithm selection.}
There have been a few attempts to design algorithm selection systems that
do not require features.
\citet{Loreggia2016} rendered problem instances as grayscale images
and classified them with a CNN.
\citet{Alissa2023} selected heuristics for online bin packing by training
recurrent networks on raw sequences of item sizes.
\citet{SalinasPinto2024} trained a custom Transformer encoder with raw
text inputs, gaining an improvement over the CNN approach but still
falling below the single best solver on industrial SAT instances.
\citet{Pellegrino2025} trained a BERT encoder from scratch on Essence
instance data for constraint programming, achieving competitive
performance for this single domain.
With a motivation that is similar to ours, \citet{Stone2024} aim at one
selector for several domains without the need for domain-specific features.
They describe instances of three problems (TSP, knapsack, graph coloring)
in a common form (text, images, graphs). They train a neural classifier on
each representation, one per domain or one for all three. Both variants
mostly perform similarly to classifiers that are built on domain-specific features.
All these approaches avoid hand-crafted features, but they train a
task-specific model on algorithm-selection labels: per domain
\citep{Loreggia2016,Alissa2023,SalinasPinto2024,Pellegrino2025} or
jointly across domains \citep{Stone2024}. Our approach avoids such
training as it uses a frozen off-the-shelf embedding model with a
non-parametric $k$-NN selector.

\paragraph{Neural and LLM-based algorithm selection.}
GraSS~\citep{Zhang2024} is a GNN-based SAT solver selector that represents
CNF instances as tripartite literal-clause graphs with hand-designed node
features and achieves competitive results on SAT benchmarks; it is
inherently domain-specific with CNF parsing and expert-designed graph
features and uses a trained GNN. \citet{Wu2024} use large language models (LLMs) to embed
algorithm source code and combine the resulting representations with
the hand-crafted instance features from ASlib. \citet{Gao2025}
propose a selector of neural solvers for two routing problems, TSP and
CVRP. They encode each instance with a graph attention network that is
trained on which solver performs best, so the instance representation is
learned for such graph-shaped routing instances. In contrast to ZeroFolio,
all these approaches require either domain-specific instance
representations or supervised training on runtime data.

\paragraph{Serialization for language models.}
\citet{Fatemi2024} study encoding of graph-structured data as text
for LLMs and show that the serialization format significantly determines
the model's performance. \citet{Yin2025} compare serialization strategies
for structured entity matching and conclude that random attribute ordering
provides higher robustness when the model is fine-tuned.
Our line-shuffling strategy (Section~\ref{sec:serialization}) also randomizes
the order, but for a different reason: to expose other parts of the instance
under truncation.

\section{Method} %
\label{sec:method}

\begin{figure}[t]
\centering
\includegraphics[width=\linewidth]{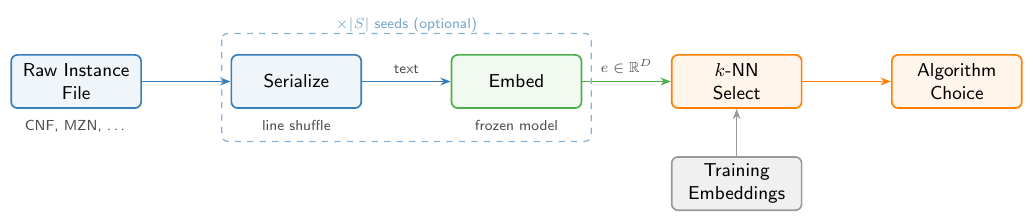}
\caption{The ZeroFolio pipeline: serialize, embed, select. The same three steps
  apply unchanged across all problem domains. Multi-seed voting
  (dashed) is optional.}
\label{fig:pipeline}
\end{figure}

We call our method \emph{ZeroFolio} since it requires zero domain
knowledge.\footnote{Supplementary material (code and data) is available at \url{https://doi.org/10.5281/zenodo.23168313}; ZeroFolio is also available as a Python package: \url{https://pypi.org/project/zerofolio/}.} It proceeds in three steps: it serializes the raw
instance file as text, embeds the text with a pretrained embedding model, and selects an
algorithm via weighted $k$-NN. See Figure~\ref{fig:pipeline} for an illustration.

\subsection{Zero Domain Knowledge Serialization}
\label{sec:serialization}

The input is the raw instance file, decompressed if required.
We consider this file as plain text without parsing or feature extraction, so we do not require any domain-specific
knowledge. The same pipeline handles CNF formulas, WCNF (MaxSAT),
QDIMACS (QBF), MiniZinc and XCSP (XML) models, answer set programs, MPS files, and
adjacency lists.
For formats with separate model and data files (e.g., MiniZinc
\texttt{.mzn} + \texttt{.dzn}), we concatenate the parts into a single string,
each part preceded by a one-line marker.

We truncate the file to a character budget $b$, where $b = 10{,}000$ unless
stated otherwise. The embedding model may truncate further according to its token
limit (Section~\ref{sec:discussion}). Since truncation discards everything outside the budget, a fixed ordering inside the file
means that later parts of the instance will not be seen. Therefore, we use
\emph{line shuffling}, where we randomly permute the lines of the file
before truncation. This way, different random seeds send different
parts of the instance to the embedding model.

Formally, let $f$ denote an instance file with lines $\ell_1, \ldots, \ell_n$.
Given a random seed $s$, the serialization is
\[
  \mathrm{serialize}(f, s) = \mathrm{truncate}(\sigma_s(\ell_1,
  \ldots, \ell_n),\; b)
\]
where $\sigma_s$ is the random permutation determined by $s$ and
$\mathrm{truncate}$ retains the first $b$ characters, cut back to the last
complete line. For formats
where line order carries no semantic meaning (e.g., CNF clauses), this
is a lossless transformation up to truncation.

\subsection{Embedding}
\label{sec:embedding}

We send the text to a pretrained embedding model via a single API call.
The model outputs a vector $e \in \mathbb{R}^D$ for a fixed model-specific
dimension $D$. Here we use the embedding model as a black box with no
fine-tuning or domain adaptation.

\subsection{Algorithm Selection via $k$-NN}
\label{sec:knn}

For a given test instance with embedding $e$, we compute a weighted score
for each algorithm $a$ from $e$'s $k$ nearest training instances using
Manhattan distance:
\[
  \mathrm{score}(a)
    = \frac{\sum_{i=1}^{k} w_i \cdot t_{i,a}}{\sum_{i=1}^{k} w_i},
  \qquad
  w_i = \frac{1}{\delta(e, e_i) + \epsilon}
\]
where $\delta$ denotes Manhattan distance, $e_i$ is the embedding of
the $i$-th neighbor, $t_{i,a}$ is the PAR10 runtime of algorithm
$a$ on that neighbor, and $\epsilon = 10^{-8}$ avoids division by zero. We then select the algorithm with the lowest weighted
score. Our ablation study (Section~\ref{sec:ablation}) shows that
this is an important design choice, especially on SAT12-ALL. Switching to a
uniform weighting degrades PAR10 by 33\% on the SAT12-ALL scenario.

\paragraph{Multi-seed voting.}
Line shuffling with different seeds results in distinct embeddings for the
same instance. We could average algorithm scores across seeds,
\[
  \mathrm{score}_{\mathrm{vote}}(a)
    = \frac{1}{|S|} \sum_{s \in S} \mathrm{score}_s(a),
\]
where $S$ is a set of random seeds and $\mathrm{score}_s(a)$ is computed
from the embedding produced with seed~$s$. Such \emph{soft voting} combines
several random truncations of the same instance. We do not use soft voting in our main
results, where we report the mean over single-seed runs. In Section~\ref{sec:seeds},
we evaluate the effect of soft voting on the SAT12-ALL scenario.

\section{Experimental Setup} %
\label{sec:setup}

\subsection{Benchmark Scenarios}
\label{sec:scenarios}

We consider 11 scenarios from the Algorithm Selection Library
(ASlib)~\citep{Bischl2016}, which span 7 domains.
Table~\ref{tab:scenarios} shows the scenarios and their main
characteristics. We start from the union of the scenarios that have been used in the 2017
Open Algorithm Selection Challenge~\citep{Lindauer2019} and the sunny-as2 evaluation~\citep{Liu2021} and keep the 9 scenarios that have a runtime metric, a text-based instance
format, publicly available instances, and at least 5 algorithms in their
portfolio. We add two older ASlib scenarios, MAXSAT12-PMS and CSP-MZN-2013.

\begin{table}[t]
\centering
\caption{The 11 ASlib benchmark scenarios used in our evaluation.
  \emph{Inst.}: number of evaluated instances.
  \emph{Algo.}: number of algorithms in the portfolio.
  \emph{Cutoff}: per-instance time limit in seconds.
  \emph{Format}: LP = logic program (ASP), SIP = pattern and target graph, MZN = MiniZinc, XML = XCSP.}
\label{tab:scenarios}
\av{\medskip}
\small
\begin{tabular}{@{}llrrrl@{}}
\toprule
Scenario & Domain & Inst. & Algo. & Cutoff (s) & Format \\
\midrule
SAT12-ALL             & SAT    & 1367 & 31 &   1200 & CNF \\
SAT03-16\_INDU        & SAT    & 1887 & 10 &   5000 & CNF \\
MAXSAT12-PMS          & MaxSAT &  875 &  6 &   2100 & WCNF \\
MAXSAT-PMS-2016       & MaxSAT &  450 & 19 &   1800 & WCNF \\
MAXSAT-WPMS-2016      & MaxSAT &  630 & 18 &   1800 & WCNF \\
QBF-2016              & QBF    &  825 & 24 &   1800 & QDIMACS \\
ASP-POTASSCO          & ASP    &  848 & 11 &    600 & LP \\
GRAPHS-2015           & Graph  & 5725 &  7 & $10^5$ & SIP \\
MIP-2016              & MIP    &  218 &  5 &   7200 & MPS \\
CSP-MZN-2013          & CSP    & 3940 & 11 &   1800 & MZN/XML \\
CSP-Minizinc-Time-2016     & CSP    &  100 & 20 &   1200 & MZN \\
\bottomrule
\end{tabular}
\end{table}

We obtained the instance files from public repositories such as SAT/MaxSAT
competition archives, QBFLIB, MIPLIB, and MiniZinc benchmarks (sources per
scenario in the supplementary material). We evaluate the instances whose files we could obtain.

\subsection{Evaluation Protocol}
\label{sec:protocol}

We use the 10-fold cross-validation (CV) splits provided by ASlib.
We use PAR10 (penalized average runtime) as the performance metric, where lower is better. It is the
runtime if the selected algorithm solves the instance within the time limit
and 10 times the time limit otherwise~\citep{Bischl2016}. The \emph{virtual best solver} (VBS) picks the best algorithm for each instance; its PAR10 is a lower bound for any selector.

We compare the performance of ZeroFolio against two baselines:
\begin{itemize}
\item The \emph{single best solver} (SBS): the algorithm with the
  lowest average PAR10 on the training instances.
\item A \emph{random forest} (RF)~\citep{Breiman2001} trained on the
  hand-crafted ASlib features, which is a standard supervised baseline for
  algorithm selection~\citep{Bischl2016}. Here we use 100 trees, median
  imputation for missing features and standard scaling. The RF
  predicts the best algorithm per instance (classification). We report the mean over ten random seeds of the forest, for this default configuration and for a per-scenario tuned variant (Section~\ref{sec:tuned-rf}).
\end{itemize}

The standard configuration for ZeroFolio uses $k{=}10$, Manhattan distance,
inverse-distance weighting, and line shuffling with a
$10{,}000$-character budget. We use the standard configuration on all
11 scenarios and do no per-scenario tuning. We chose this configuration based on
preliminary experiments, mainly on SAT12-ALL (for $k$, see Section~\ref{sec:ablation}).

\subsection{Embedding Model}
\label{sec:model}

\begin{sloppypar}
As our primary embedding model, we use \texttt{gemini-embedding-2-preview} (Gemini~2 below), a
proprietary model by Google with 3072 output dimensions. In addition, we compare three
alternative embedding models: the Gemini predecessor model \texttt{gemini-embedding-001} (Gemini~1; 3072 dimensions),
OpenAI's \texttt{text-embedding-3-large} (3072 dimensions, 8k token context),
and the open-source \texttt{qwen3-embedding-8b}
(4096 dimensions, 32k token context).
\end{sloppypar} We access Gemini~2 via the Vertex AI API or OpenRouter and all other models
via OpenRouter. All models are used without fine-tuning. See Section~\ref{sec:model-comparison}
for the model comparison.

\section{Results} %
\label{sec:results}

\subsection{Main Results}
\label{sec:main-results}

We summarize the results in Table~\ref{tab:main}. We use the same
configuration for all scenarios ($k{=}10$, Manhattan distance,
inverse-distance weighting). Since line shuffling is randomized, we
report ZeroFolio as the mean PAR10 over the serialization seeds. We
analyze the variance in Section~\ref{sec:seeds}. ZeroFolio outperforms
the RF in 9 of the 11 scenarios, in 8 of them with every single seed. The differences are
often large. On ASP-POTASSCO, ZeroFolio obtains 570 where the RF
obtains 784. On QBF-2016 it is 2124 versus 2494, and on the two CSP
scenarios the difference is at least $16\%$. On GRAPHS-2015 the RF (31,028) is worse
than even the SBS (29,155), whereas ZeroFolio (26,385) closes a third of
the SBS--VBS gap.

The two exceptions are SAT12-ALL and MAXSAT12-PMS. On SAT12-ALL, the RF
with 956 clearly beats ZeroFolio with mean 1158. No seed crosses below
the RF. On MAXSAT12-PMS the two methods are
effectively tied (mean 3756 vs.\ 3729). With a Wilcoxon signed-rank test across the 10 CV folds, we find
ZeroFolio's advantage significant ($p{<}0.05$) with seed 42 on 5 of the 9 scenarios
that it wins on average, and on 4 with each of the other two seeds. The default RF depends on its random
seed, so Table~\ref{tab:main} reports its mean over ten seeds. Single forest seeds range from 918 to 996
on SAT12-ALL and from 2627 to 2978 on MIP-2016, and ZeroFolio beats the RF on 9 or 10
of the 11 scenarios under every forest seed.

On the three scenarios with published AutoFolio PAR10 from the 2015
ICON Challenge on Algorithm Selection (Table~\ref{tab:main}, AF$^{\star}$), AutoFolio has the
lower PAR10. However, AutoFolio is configured per scenario, whereas
ZeroFolio without any per-scenario configuration stays close, within $6$--$9\%$
(570 vs.\ 525 on ASP-POTASSCO, 3756 vs.\ 3559 on MAXSAT12-PMS, and
1158 vs.\ 1066 on SAT12-ALL). The comparison is not exact: the competition
used the full scenarios and 10 bootstrap train/test splits instead of the
ASlib folds, allowed pre-solvers, and counted feature costs.

\begin{table}[t]
\centering
\caption{Main results: PAR10 across 11 ASlib scenarios. ZF: ZeroFolio
  ($k$-NN on embeddings, $k{=}10$), reported as the \emph{mean over
  serialization seeds} (per-seed variation in Section~\ref{sec:seeds}).
  RF: default random forest, mean over ten random seeds.
  RF$_t$: random forest with per-scenario hyperparameters selected by
  nested cross-validation, mean over the same ten seeds. AF$^{\star}$: AutoFolio PAR10 from the 2015
  ICON Challenge~\citep[Table~D.6; 10 bootstrap train/test splits]{Lindauer2019};
  ``--'' indicates no published number. AF$^{\star}$ uses the full
  scenario (1294 and 1614 instances for ASP-POTASSCO and SAT12-ALL),
  whereas our evaluation uses the subset with available raw files (848
  and 1367); MAXSAT12-PMS counts are nearly identical (876 vs.\ 875).
  Bold indicates the best among RF, RF$_t$, and ZF per scenario (for Gap\%, the better of RF and ZF).
  Gap\%: percentage of SBS--VBS gap closed.}
\label{tab:main}
\av{\medskip}
\small
\begin{tabular}{@{}lrrrrrrrr@{}}
\toprule
 & & & & & & & \multicolumn{2}{c}{Gap\%} \\
\cmidrule(lr){8-9}
Scenario & SBS & RF & RF$_t$ & AF$^{\star}$ & ZF & VBS & RF & ZF \\
\midrule
SAT12-ALL          & 3066 & 956 & \textbf{952} & 1066 & 1158 &  271 & \textbf{75} & 68 \\
SAT03-16\_INDU     & 10097 & 9504 & 9500 & -- & \textbf{9292} & 7152 & 20 & \textbf{27} \\
MAXSAT12-PMS       & 4899 & \textbf{3729} & 3733 & 3559 & 3756 & 3131 & \textbf{66} & 65 \\
MAXSAT-PMS-2016    & 2965 & 3203 & 3211 & -- & \textbf{2338} & 1833 & $-$21 & \textbf{55} \\
MAXSAT-WPMS-2016   & 3893 & 3590 & 3526 & -- & \textbf{3379} & 2630 & 24 & \textbf{41} \\
QBF-2016           & 3667 & 2494 & 2509 & -- & \textbf{2124} & 1209 & 48 & \textbf{63} \\
ASP-POTASSCO       & 1015 & 784 & 796 & 525 & \textbf{570} &  440 & 40 & \textbf{77} \\
GRAPHS-2015        & 29155 & 31028 & 31339 & -- & \textbf{26385} & 20769 & $-$22 & \textbf{33} \\
MIP-2016           & 3008 & 2860 & \textbf{2214} & -- & 2516 &  282 & 5 & \textbf{18} \\
CSP-MZN-2013       & 9461 & 5394 & 5391 & -- & \textbf{4529} & 3950 & 74 & \textbf{89} \\
CSP-Minizinc-Time-2016  & 3612 & 2944 & 3041 & -- & \textbf{2462} & 2062 & 43 & \textbf{74} \\
\bottomrule
\end{tabular}
\end{table}

\subsection{Seed Sensitivity and Voting}
\label{sec:seeds}

Because line shuffling is randomized, the embedding of an instance
depends on a seed. The means reported in Table~\ref{tab:main} average
over seeds. To see how much the result depends on the seed, we embed every
scenario with three different seeds (nine for SAT12-ALL) and report the
standard deviation of PAR10 across the seeds (for every scenario in the supplementary material). On the
nine scenarios where ZeroFolio wins on average, it wins on every seed except on
MIP-2016. The standard deviation is often small compared to the ZeroFolio--RF
margin, e.g., $\pm 28$ on
ASP-POTASSCO (mean 570 vs.\ RF 784) and $\pm 6$ on CSP-MZN-2013.
MIP-2016 is the only scenario with high variance ($\pm 660$). There, the worst seed (3252)
loses to the RF (2860).

On SAT12-ALL, across nine seeds, single-seed PAR10 ranges from 1108 to 1222 (mean
$1158 \pm 37$ at $k{=}10$, $1077 \pm 55$ at $k{=}5$); no seed beats the
RF (956) at either $k{=}10$ or $k{=}5$. We also
tested the multi-seed soft voting of Section~\ref{sec:knn}. Over all 36
two-seed pairs ($k{=}5$) it yields mean PAR10 $1030 \pm 55$; the best
pair reaches 893, and 3 of the 36 pairs (7 of the 84 triples) fall
below the RF, but the average pair does not. Voting reduces the mean
PAR10 from 1077 to 1030 but leaves the standard deviation unchanged.
However, it does not turn SAT12-ALL into a win on average.

\subsection{Tuned Random-Forest Baseline}
\label{sec:tuned-rf}

To check that ZeroFolio is not merely beating an under-tuned baseline,
we strengthen the random forest (column RF$_t$ in
Table~\ref{tab:main}). For each scenario we select the RF
hyperparameters (maximum depth, maximum features, minimum leaf size) by
an inner 3-fold cross-validation on each training fold, scored by PAR10,
and evaluate the chosen model on the held-out fold. The hyperparameters are chosen
using the training folds only, so the held-out fold on which we make the
evaluation does not influence the choice. Tuning helps substantially only on MIP-2016
(2860$\to$2214), where the tuned RF becomes better than ZeroFolio (2516).
Elsewhere it changes PAR10 by less than 4\%, and on six scenarios it even makes the RF perform worse. Even if we take the better of the
default and tuned forest per scenario, ZeroFolio wins on 8 of the 11.

\subsection{Ablation Study}
\label{sec:ablation}

We carry out an ablation study on SAT12-ALL to identify the
contribution of each design choice. We use the standard
configuration ($k{=}10$, Manhattan distance, inverse-distance
weighting, line shuffle) but vary the setting of one dimension at a
time. We summarize the results in Table~\ref{tab:ablation}.

We see that inverse-distance weighting has a large impact on the result.
If we switch to uniform weighting, PAR10 increases from 1122 to 1497
($+33$\%). Also, line shuffling has an impact. The raw
file order yields 1217 versus 1122 for shuffling ($+8$\%; $+5$\% against the mean over nine seeds).
The distance metric has relatively little impact. On this scenario, cosine is even stronger than Manhattan distance
(1098 vs.\ 1122). The naive baseline (raw text, cosine, uniform) reaches only 1495.

On SAT12-ALL, $k{=}5$ is better than $k{=}10$ (1077 vs.\ 1158, mean over nine seeds). We fixed $k{=}10$ after preliminary runs on the other scenarios. With a single seed, every tested $k$ from 7 to 20 beats the RF on 9 or 10 of the 11 scenarios, and $k{=}5$ on 8 (supplementary material).

To test whether pretrained embeddings are necessary, we replace them
with TF-IDF character $n$-gram features of the same dimensionality
(3072), fitted on the training folds and computed from the unshuffled first 10,000 characters, and apply the same $k$-NN selector. With character 2--4-grams, TF-IDF
reaches 1527 on SAT12-ALL (1538 with 3--5-grams). This is better than SBS (3066) but
substantially worse than Gemini embeddings of the same unshuffled text (1217). This gap suggests
that pretrained representations capture information beyond
surface-level text statistics.

To verify that our design choices generalize, we repeat the distance and weighting
ablations on three additional scenarios
(Table~\ref{tab:cross-ablation}). We see that inverse-distance weighting remains
important on QBF-2016, but has little impact on ASP-POTASSCO and
CSP-MZN-2013. The large gain on SAT12-ALL originates from duplicate instances,
which are embedded at distance zero: without them, uniform weights are only 2\% worse. With all duplicates removed, ZeroFolio (seed 42) still beats the RF on the same 10 of 11 scenarios as on all instances (Table~\ref{tab:models}; supplementary material). Uniform weights beat $1/\delta$ on 5 of the
11 scenarios (supplementary material). Manhattan and cosine distance achieve nearly identical results on these
three scenarios. In none of the three does a variant beat the standard configuration by more
than 2.5\%. Hence we keep it fixed for all scenarios.

\begin{table}[t]
\centering
\caption{Ablation on SAT12-ALL. Standard configuration: $k{=}10$,
  Manhattan, $1/\delta$ weighting, line shuffle, single seed. Each row
  varies one dimension from the standard (the naive baseline varies three).}
\label{tab:ablation}
\medskip
\small
\begin{tabular}{@{}llr@{}}
\toprule
Dimension & Variant & PAR10 \\
\midrule
\multicolumn{2}{@{}l}{Standard configuration} & 1122 \\
\multicolumn{2}{@{}l}{Naive baseline (raw + cosine + uniform)} & 1495 \\
\midrule
Serialization & raw (no shuffle) & 1217 \\
Distance      & cosine           & 1098 \\
Weighting     & uniform          & 1497 \\
$k$           & $k{=}5$          & 1010 \\
$k$           & $k{=}20$         & 1176 \\
\bottomrule
\end{tabular}
\end{table}

\begin{table}[t]
\centering
\caption{Cross-domain ablation ($k{=}10$, single seed). Each cell shows
  PAR10 when one dimension departs from the standard configuration (Manhattan,
  $1/\delta$). $\Delta$: relative change from standard.}
\label{tab:cross-ablation}
\medskip
\small
\begin{tabular}{@{}lrrrrr@{}}
\toprule
 & & \multicolumn{2}{c}{Cosine} & \multicolumn{2}{c}{Uniform wt.} \\
\cmidrule(lr){3-4}\cmidrule(lr){5-6}
Scenario & Std & PAR10 & $\Delta$ & PAR10 & $\Delta$ \\
\midrule
SAT12-ALL     & 1122 & 1098 & $-2\%$ & 1497 & $+33\%$ \\
QBF-2016      & 1979 & 1959 & $-1\%$ & 2220 & $+12\%$ \\
ASP-POTASSCO  &  539 &  539 & $ 0\%$ &  526 & $-2\%$  \\
CSP-MZN-2013  & 4524 & 4524 & $ 0\%$ & 4528 & $ 0\%$  \\
\bottomrule
\end{tabular}
\end{table}

\subsection{Embedding Model Comparison}
\label{sec:model-comparison}

In Table~\ref{tab:models}, we compare four embedding models at seed 42. Gemini~2 (proprietary, 3072 dimensions)
outperforms the RF baseline on 10 of 11 scenarios. Its predecessor
Gemini~1 (same dimensions) beats RF on 9 of 11, and OpenAI text-embedding-3-large (also 3072
dimensions) on 8 of 11.
The open-source Qwen3-Embedding-8B (4096 dimensions, 32k
context) beats RF on 6 of 11.

Gemini~2, Gemini~1, and OpenAI all produce vectors with 3072 dimensions,
so their differences cannot come from the dimension. They can come from
training data and architecture, and for Gemini~1 also from its shorter
input (2048 instead of 8192 tokens). Differences smaller than the
seed-to-seed variation of Section~\ref{sec:seeds} can be neglected.

\begin{table}[t]
\centering
\caption{Embedding model comparison ($k{=}10$, Manhattan, $1/\delta$, serialization
  seed 42 for every model). Gem.\,2 and Gem.\,1: Google, 3072d. OAI: OpenAI, 3072d.
  Qwen3: open-source, 4096d. The ZF column of Table~\ref{tab:main} is instead a mean
  over seeds; with it, Gem.\,2 beats RF on 9 of 11. Bold: best embedding result per scenario (single seed).}
\label{tab:models}
\av{\medskip}
\small
\begin{tabular}{@{}lrrrrr@{}}
\toprule
Scenario & RF & Gem.\,2 & Gem.\,1 & OAI & Qwen3 \\
\midrule
SAT12-ALL          & 956 & \textbf{1122} & 1126 & 1226 & 1537 \\
SAT03-16\_INDU     & 9504 & \textbf{9270} & 9318 & 9282 & 9709 \\
MAXSAT12-PMS       & 3729 & 3684 & 3709 & \textbf{3635} & 3806 \\
MAXSAT-PMS-2016    & 3203 & \textbf{2319} & 2324 & 2478 & 2640 \\
MAXSAT-WPMS-2016   & 3590 & \textbf{3378} & 3457 & 3542 & 3859 \\
QBF-2016           & 2494 & \textbf{1979} & 2217 & 2566 & 3178 \\
ASP-POTASSCO       & 784 & \textbf{539} & 574 & 576 & 646 \\
GRAPHS-2015        & 31028 & 26727 & 26014 & \textbf{25881} & 26256 \\
MIP-2016           & 2860 & 2318 & 3313 & 2981 & \textbf{2002} \\
CSP-MZN-2013       & 5394 & 4524 & 4537 & \textbf{4511} & 4680 \\
CSP-Minizinc-Time-2016  & 2944 & 2309 & 2536 & \textbf{2188} & 2316 \\
\midrule
Beats RF           & -- & 10/11 & 9/11 & 8/11 & 6/11 \\\bottomrule
\end{tabular}
\end{table}

\subsection{Hybrid Selection}
\label{sec:hybrid}

We combine ZeroFolio with each of the ten default forests (on ASlib features) via soft voting and report the mean over the forests.
For each algorithm, we take the inverse of the $k$-NN score and the RF's class
probability, normalize each to sum to one over the algorithms, and average
the two with equal weight ($\alpha{=}0.5$).

On SAT12-ALL, where the random forest is clearly the stronger single
selector, the hybrid method achieves PAR10 $857 \pm 23$ over the nine seeds
and beats ZeroFolio (1158) and both forests (956 and 952) on every seed. Consequently, the two representations are complementary
on this scenario. Using min-max normalization, the hybrid reaches $894 \pm 15$
and also beats both forests on every seed. Where ZeroFolio clearly
dominates, the hybrid does not help: on ASP-POTASSCO, QBF-2016, and CSP-MZN-2013
the hybrid (seed 42) beats both forests but is behind ZeroFolio alone. Joining the two
vectors into one long vector and running $k$-NN on it performs worse than soft
voting. In the joint vector, the 3072 embedding coordinates outnumber the 115 standardized
features and dominate the distance, so the features have little impact.

\section{Discussion} %
\label{sec:discussion}

\begin{figure}[t]
\centering
\includegraphics[width=0.85\linewidth]{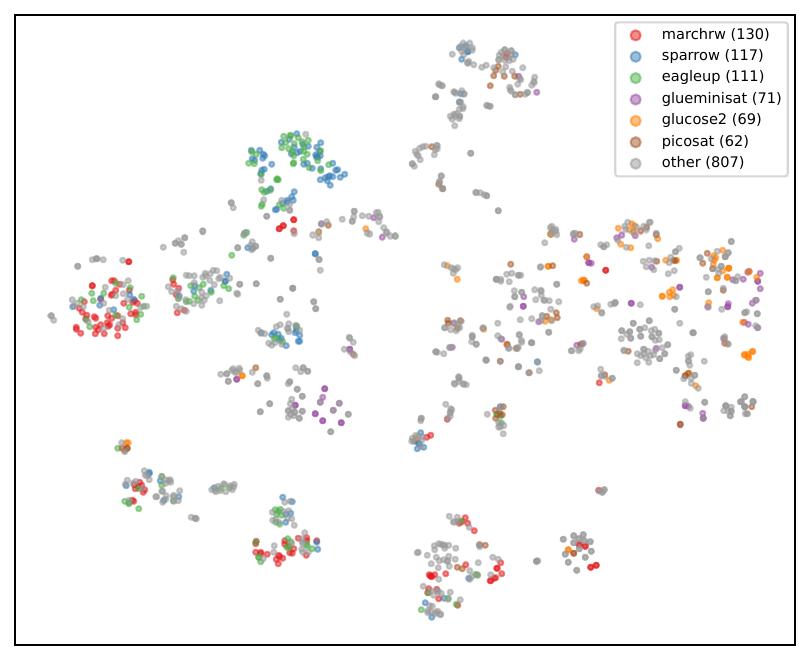}
\caption{t-SNE projection of the Gemini~2 embeddings (serialization seed 42) of the 1367
  SAT12-ALL instances. Colors mark the six most frequent best solvers, with instance counts
  in the legend; \emph{other} collects the instances with any other best solver. Instances with the
  same best solver partly cluster.}
\label{fig:tsne}
\end{figure}

\paragraph{Why does it work?}
Pretrained text embedding models are typically trained on a broad set of text
collected from the Internet, including natural-language text, source code, configuration
files, and structured data. A CNF formula in DIMACS format is not
natural language, but it is structured text with certain
regularities (token distributions, clause length patterns, comment
headers).
Our results indicate that similarities found by embedding models are sufficient to
identify instances with similar solver performance.\av{ A t-SNE projection
of the SAT12-ALL embeddings (Figure~\ref{fig:tsne}) shows that instances with the same best solver partly cluster, which suggests that the embedding space captures instance
structure that is relevant for solver choice.}

\paragraph{Why a pretrained model?}
The TF-IDF baseline of Section~\ref{sec:ablation} replaces the pretrained model by simple text statistics of the same raw text. On SAT12-ALL, it
reaches PAR10 $1527$, far worse than both the pretrained
embedding ($1217$) and the RF ($956$). Counting which short character sequences occur in the file, as TF-IDF does,
is not enough.

\paragraph{Why $k$-NN?}
We deliberately chose a non-parametric selector, since a random forest trained on
the same $3072$-dimensional embeddings scores only $1513$ on SAT12-ALL,
far worse than $k$-NN ($1122$). A plausible explanation is that tree ensembles split on individual
coordinates and degrade on dense, high-dimensional representations,
whereas $k$-NN makes use of the global topology of
the embedding space.

\paragraph{What ``feature-free'' means.}
We do not just outsource the domain knowledge to the embedding model.
Since the model was not trained on the algorithm-selection task, its knowledge comes only from
generic web-scale training that measures similarities of text. It was not trained on
whether text represents instances that are suitable for one solver or another.

\paragraph{Scope.}
Our method can be applied to any problem domain with text-based instance
formats. Binary or image-based formats require a different
serialization step, but could possibly be adapted in future work. Within our scope, the same pipeline handles all
considered instance formats without modification. No domain expertise is
required. In contrast to hand-crafted feature extractors, our serialization
reads the file as plain text without parsing. However, like all supervised selection
approaches, we still require labeled runtime data per
scenario.

\paragraph{Truncation and the character budget.}
Truncating to a tight character budget could destroy structural
continuity. However, we limit this effect, and it appears small.
First, line shuffling makes the retained window a random sample of the
whole file, not just a prefix, so the embedding sees material
from the entire instance; a fixed prefix is $5\%$ worse
than the mean over nine seeds. Second, the performance does not indicate a trend in the budget.
On SAT12-ALL, the budgets of 5000, 7000, 10,000, and
30,000 characters yield PAR10 of $997$, $1117$, $1010$, and $1010$, respectively ($k{=}5$),
a range that is within the seed-to-seed variation (Section~\ref{sec:seeds}).
The identical values for 10,000 and 30,000 characters reflect the 8192-token input
limit of Gemini~2, which on SAT12-ALL is reached between 7000 and 10,000
characters. This suggests that the selector depends on coarse instance
statistics that survive truncation.

\paragraph{Limitations.}
Our best results are achieved on proprietary embedding models (Gemini,
OpenAI), with open-source alternatives mostly behind
(Section~\ref{sec:model-comparison}). The proprietary models involve API costs, where a single
embedding call with Gemini is about 0.2 cents. This cost is small compared to the cost of running solvers,
but a reliance on proprietary embedding models also introduces availability and
reproducibility risks. The SAT12-ALL scenario is one where the
feature-based approach is genuinely stronger than ZeroFolio. On this scenario, AutoFolio scores 1066 in the
2015 ICON Challenge, and our own tuned random forest reaches 952, both
below ZeroFolio's 1158. The RF is also not the
strongest feature-based selector. Our standard $k$-NN selector on standardized
ASlib features beats the RF on 10 of 11 scenarios, so part of ZeroFolio's margin over
the RF results from the selector, not the representation. It even beats ZeroFolio on 8 of 11
(supplementary material). Finally,
whether task-specific fine-tuning would help, at the cost of the
zero-domain-knowledge framing, is outside our scope.

\paragraph{Feature computation cost.}
Our approach also avoids the costs and failure modes of feature
extraction. Recall that the original SATzilla feature extractor fails on
over a fifth of modern SAT competition instances (Section~\ref{sec:intro}). Feature computation can take minutes per
instance~\citep{Shavit2024}. Our pipeline (serialize, embed, select) needs one embedding API call per
instance; serialization takes milliseconds for most of our files, and
$k$-NN retrieval is negligible. For domains where no established feature
extractor exists, our method could offer an immediate off-the-shelf
alternative.

\section{Conclusion} %
\label{sec:conclusion}

We present ZeroFolio, an approach to algorithm selection based on
pretrained text embeddings that does not require any feature
engineering or task-specific training.

Our experiments show that ZeroFolio outperforms a random forest trained
on hand-crafted features on 9 of 11 ASlib scenarios over 7 domains; it
nearly ties on one scenario and loses on another. In 8 of the 9 scenarios it
wins with every serialization seed. On the three scenarios with published
AutoFolio competition results, ZeroFolio comes within 6--9\% of AutoFolio,
which is tuned per scenario. An ablation study on the
SAT12-ALL scenario shows that the inverse-distance weighting and line shuffling contribute
to the good performance of the system. ZeroFolio can be applied unchanged to any domain
with text-based instance formats.

Beyond algorithm selection, our embedding approach could be useful for
benchmark design~\citep{SmithMiles2015}: from a large pool of instances,
one can select a subset that covers the space evenly, or indicate
underrepresented regions.

\begin{acknowledgements}
\cv{\begin{sloppypar}Research was supported by the Austrian Science Fund (FWF) within the projects 10.55776/COE12 and 10.55776/P36420.\end{sloppypar}}
\av{\begin{sloppypar}Research was supported by the Austrian Science Fund (FWF) within the projects 10.55776/COE12 and 10.55776/P36420.\end{sloppypar}}
\end{acknowledgements}

\bibliography{references}

\begin{thebibliography}{}

\bibitem[Alissa et~al., 2023]{Alissa2023}
Alissa, M., Sim, K., and Hart, E. (2023).
\newblock Automated algorithm selection: from feature-based to feature-free
  approaches.
\newblock {\em J. Heuristics}, 29(1):1--38.

\bibitem[Bischl et~al., 2016]{Bischl2016}
Bischl, B., Kerschke, P., Kotthoff, L., Lindauer, M., Malitsky, Y.,
  Fr{\'{e}}chette, A., Hoos, H.~H., Hutter, F., Leyton{-}Brown, K., Tierney,
  K., and Vanschoren, J. (2016).
\newblock {ASlib}: {A} benchmark library for algorithm selection.
\newblock {\em Artif. Intell.}, 237:41--58.

\bibitem[Breiman, 2001]{Breiman2001}
Breiman, L. (2001).
\newblock Random forests.
\newblock {\em Mach. Learn.}, 45(1):5--32.

\bibitem[Fatemi et~al., 2024]{Fatemi2024}
Fatemi, B., Halcrow, J., and Perozzi, B. (2024).
\newblock Talk like a graph: Encoding graphs for large language models.
\newblock In {\em The Twelfth International Conference on Learning
  Representations, {ICLR} 2024, Vienna, Austria, May 7-11, 2024}.
  OpenReview.net.

\bibitem[Gao et~al., 2025]{Gao2025}
Gao, C., Shang, H., Xue, K., and Qian, C. (2025).
\newblock Neural solver selection for combinatorial optimization.
\newblock In Singh, A., Fazel, M., Hsu, D., Lacoste{-}Julien, S., Berkenkamp,
  F., Maharaj, T., Wagstaff, K., and Zhu, J., editors, {\em Forty-second
  International Conference on Machine Learning, {ICML} 2025, Vancouver, BC,
  Canada, July 13-19, 2025}, volume 267 of {\em Proceedings of Machine Learning
  Research}. {PMLR} / OpenReview.net.

\bibitem[Kotthoff, 2016]{Kotthoff2016}
Kotthoff, L. (2016).
\newblock Algorithm selection for combinatorial search problems: {A} survey.
\newblock In Bessiere, C., {De Raedt}, L., Kotthoff, L., Nijssen, S.,
  O'Sullivan, B., and Pedreschi, D., editors, {\em Data Mining and Constraint
  Programming - Foundations of a Cross-Disciplinary Approach}, volume 10101 of
  {\em Lecture Notes in Computer Science}, pages 149--190. Springer.

\bibitem[Lindauer et~al., 2015]{Lindauer2015}
Lindauer, M., Hoos, H.~H., Hutter, F., and Schaub, T. (2015).
\newblock {AutoFolio}: An automatically configured algorithm selector.
\newblock {\em J. Artif. Intell. Res.}, 53:745--778.

\bibitem[Lindauer et~al., 2019]{Lindauer2019}
Lindauer, M., van Rijn, J.~N., and Kotthoff, L. (2019).
\newblock The algorithm selection competitions 2015 and 2017.
\newblock {\em Artif. Intell.}, 272:86--100.

\bibitem[Liu et~al., 2021]{Liu2021}
Liu, T., Amadini, R., Gabbrielli, M., and Mauro, J. (2021).
\newblock sunny-as2: Enhancing {SUNNY} for algorithm selection.
\newblock {\em J. Artif. Intell. Res.}, 72:329--376.

\bibitem[Loreggia et~al., 2016]{Loreggia2016}
Loreggia, A., Malitsky, Y., Samulowitz, H., and Saraswat, V.~A. (2016).
\newblock Deep learning for algorithm portfolios.
\newblock In Schuurmans, D. and Wellman, M.~P., editors, {\em Proceedings of
  the Thirtieth {AAAI} Conference on Artificial Intelligence, February 12-17,
  2016, Phoenix, Arizona, {USA}}, pages 1280--1286. {AAAI} Press.

\bibitem[Nudelman et~al., 2004]{Nudelman2004}
Nudelman, E., Leyton{-}Brown, K., Hoos, H.~H., Devkar, A., and Shoham, Y.
  (2004).
\newblock Understanding random {SAT:} beyond the clauses-to-variables ratio.
\newblock In Wallace, M., editor, {\em Principles and Practice of Constraint
  Programming - {CP} 2004, 10th International Conference, {CP} 2004, Toronto,
  Canada, September 27 - October 1, 2004, Proceedings}, volume 3258 of {\em
  Lecture Notes in Computer Science}, pages 438--452. Springer.

\bibitem[Pellegrino et~al., 2025]{Pellegrino2025}
Pellegrino, A., Akg{\"{u}}n, {\"{O}}., Dang, N., Kiziltan, Z., and Miguel, I.
  (2025).
\newblock Transformer-based feature learning for algorithm selection in
  combinatorial optimisation.
\newblock In {Garcia de la Banda}, M., editor, {\em 31st International
  Conference on Principles and Practice of Constraint Programming, {CP} 2025,
  Glasgow, Scotland, August 10-15, 2025}, volume 340 of {\em LIPIcs}, pages
  31:1--31:22. Schloss Dagstuhl - Leibniz-Zentrum f{\"{u}}r Informatik.

\bibitem[Rice, 1976]{Rice1976}
Rice, J.~R. (1976).
\newblock The algorithm selection problem.
\newblock In {\em Advances in Computers}, volume~15, pages 65--118. Elsevier.

\bibitem[Salinas{-}Pinto et~al., 2024]{SalinasPinto2024}
Salinas{-}Pinto, A., Alvarado{-}Ulloa, B., Hochbaum, D.~S.,
  Francia{-}Carrami{\~{n}}ana, M., {\~{N}}anculef, R., and {As{\'\i}n
  Ach{\'a}}, R.~J. (2024).
\newblock Text-based feature-free automatic algorithm selection.
\newblock In Coenen, F., Fred, A., and Bernardino, J., editors, {\em
  Proceedings of the 16th International Joint Conference on Knowledge
  Discovery, Knowledge Engineering and Knowledge Management, {IC3K} 2024,
  Volume 1: KDIR, Porto, Portugal, November 17-19, 2024}, pages 267--274.
  {SCITEPRESS}.

\bibitem[Shavit and Hoos, 2024]{Shavit2024}
Shavit, H. and Hoos, H.~H. (2024).
\newblock Revisiting {SATZilla} features in 2024.
\newblock In Chakraborty, S. and Jiang, J.~R., editors, {\em 27th International
  Conference on Theory and Applications of Satisfiability Testing, {SAT} 2024,
  Pune, India, August 21-24, 2024}, volume 305 of {\em LIPIcs}, pages
  27:1--27:26. Schloss Dagstuhl - Leibniz-Zentrum f{\"{u}}r Informatik.

\bibitem[Smith{-}Miles and Bowly, 2015]{SmithMiles2015}
Smith{-}Miles, K. and Bowly, S. (2015).
\newblock Generating new test instances by evolving in instance space.
\newblock {\em Comput. Oper. Res.}, 63:102--113.

\bibitem[Stone et~al., 2024]{Stone2024}
Stone, C., Renau, Q., Miguel, I., and Hart, E. (2024).
\newblock An evaluation of domain-agnostic representations to enable multi-task
  learning in combinatorial optimisation.
\newblock In {\em Learning and Intelligent Optimization (LION 18)}, volume
  14990 of {\em Lecture Notes in Computer Science}, pages 399--414. Springer.

\bibitem[Wu et~al., 2024]{Wu2024}
Wu, X., Zhong, Y., Wu, J., Jiang, B., and Tan, K.~C. (2024).
\newblock Large language model-enhanced algorithm selection: Towards
  comprehensive algorithm representation.
\newblock In {\em Proceedings of the Thirty-Third International Joint
  Conference on Artificial Intelligence, {IJCAI} 2024, Jeju, South Korea,
  August 3-9, 2024}, pages 5235--5244. ijcai.org.

\bibitem[Xu et~al., 2008]{Xu2008}
Xu, L., Hutter, F., Hoos, H.~H., and Leyton{-}Brown, K. (2008).
\newblock {SATzilla}: Portfolio-based algorithm selection for {SAT}.
\newblock {\em J. Artif. Intell. Res.}, 32:565--606.

\bibitem[Yin et~al., 2025]{Yin2025}
Yin, H., Kim, J., Mathur, P., Sarker, K., and Bansal, V. (2025).
\newblock How to talk to language models: Serialization strategies for
  structured entity matching.
\newblock In Chiruzzo, L., Ritter, A., and Wang, L., editors, {\em Findings of
  the Association for Computational Linguistics: {NAACL} 2025, Albuquerque, New
  Mexico, USA, April 29 - May 4, 2025}, volume {NAACL} 2025 of {\em Findings of
  {ACL}}, pages 7836--7850. Association for Computational Linguistics.

\bibitem[Zhang et~al., 2024]{Zhang2024}
Zhang, Z., Ch{\'{e}}telat, D., Cotnareanu, J., Ghose, A., Xiao, W., Zhen, H.,
  Zhang, Y., Hao, J., Coates, M., and Yuan, M. (2024).
\newblock {GraSS}: Combining graph neural networks with expert knowledge for
  {SAT} solver selection.
\newblock In Baeza{-}Yates, R. and Bonchi, F., editors, {\em Proceedings of the
  30th {ACM} {SIGKDD} Conference on Knowledge Discovery and Data Mining, {KDD}
  2024, Barcelona, Spain, August 25-29, 2024}, pages 6301--6311. {ACM}.

\end{thebibliography}

\end{document}